\documentclass{article}

\usepackage{microtype}
\usepackage{graphicx}
\usepackage{subcaption}
\usepackage{booktabs} % for professional tables

\usepackage{hyperref}
\usepackage{csquotes}

 \usepackage[preprint]{icml2026}

\usepackage{amsmath}
\usepackage{amssymb}
\usepackage{mathtools}
\usepackage{amsthm}
\usepackage{tcolorbox}
\usepackage{multirow}

\usepackage[capitalize,noabbrev]{cleveref}

\theoremstyle{plain}

\theoremstyle{definition}

\theoremstyle{remark}

\usepackage[textsize=tiny]{todonotes}

\icmltitlerunning{MDER-DR: Multi‑Hop Question Answering with Entity‑Centric Summaries}

\begin{document}

\twocolumn[
  \icmltitle{MDER-DR: Multi‑Hop Question Answering with Entity‑Centric Summaries}



  \icmlsetsymbol{equal}{*}

  \begin{icmlauthorlist}
    \icmlauthor{Riccardo Campi}{yyy}
    \icmlauthor{Nicolò Oreste Pinciroli Vago}{yyy}
    \icmlauthor{Mathyas Giudici}{yyy}
    \icmlauthor{Marco Brambilla}{yyy}
    \icmlauthor{Piero Fraternali}{yyy}
  \end{icmlauthorlist}

  \icmlaffiliation{yyy}{Department of Electronics, Information and Bioengineering, Politecnico di Milano, Milan, Italy}

  \icmlcorrespondingauthor{Riccardo Campi}{riccardo.campi@polimi.it}

  \icmlkeywords{Machine Learning, ICML}

  \vskip 0.3in
]



\printAffiliationsAndNotice{}  % no special notice (required even if empty)

\begin{abstract}
  Retrieval-Augmented Generation (RAG) over Knowledge Graphs (KGs) suffers from the fact that indexing approaches may lose important contextual nuance when text is reduced to triples, thereby degrading performance in downstream Question-Answering (QA) tasks, particularly for multi-hop QA, which requires composing answers from multiple entities, facts, or relations.  
  We propose a domain-agnostic, KG-based QA framework that covers both the indexing and retrieval/inference phases.  A new indexing approach called {Map‑Disambiguate‑Enrich‑Reduce (MDER)} generates context‑derived triple descriptions and subsequently integrates them with entity‑level summaries, thus avoiding the need for explicit traversal of edges in the graph during the QA retrieval phase.
  Complementing this, we introduce {Decompose‑Resolve (DR)}, a retrieval mechanism that decomposes user queries into resolvable triples and grounds them in the KG via iterative reasoning.
  Together, MDER and DR form an LLM-driven QA pipeline that is robust to sparse, incomplete, and complex relational data.
  Experiments show that on standard and domain-specific benchmarks, MDER-DR achieves substantial improvements over standard RAG baselines (up to 66\%), while maintaining cross-lingual robustness.
  Our code is available at \url{https://github.com/DataSciencePolimi/MDER-DR\_RAG}.
\end{abstract}

%%%%%%%%%%%%%%%%%%%%%%%%%%
% INTRODUCTION
%%%%%%%%%%%%%%%%%%%%%%%%%%

\section{Introduction}
\label{sec:introduction}

\begin{figure}[t]
    \centering
    \includegraphics[width=\linewidth]{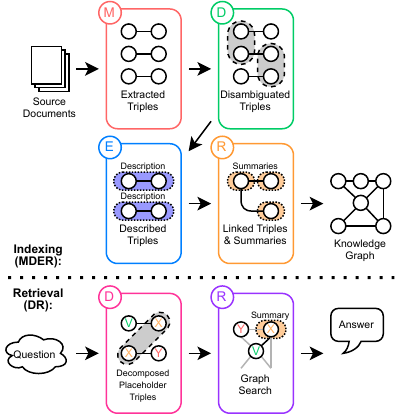}
    \caption{High‑level overview of our KG-based question answering framework, featuring the MDER indexing approach for condensing multi‑hop information during KG construction so that downstream retrieval does not need explicit hop traversal, and the DR retrieval mechanism for resolving queries through structured graph reasoning using MDER summaries, allowing multi‑entity reasoning without explicit KG path traversal.}
    \label{fig:MDER-DR}
\end{figure}

Knowledge Graphs (KGs) are central for structuring information, as they represent facts as triples in the form of subject, predicate, and object, that can be queried, reasoned over, and extended~\cite{KGs_aidan}.
Furthermore, the advancement of Large Language Models (LLMs) has accelerated both the automatic construction of KGs from text, improving their indexing performance, and their use as external knowledge sources for Question-Answering (QA) tasks, reinforcing their role in modern information retrieval \cite{LLMs_and_KGs}, which is typically referred to as graph-based Retrieval-Augmented Generation (RAG) \cite{rag_survey}.

However, extracting triples from unstructured text is inherently lossy \cite{automatic_kg_construction}. Important contextual nuances, such as the scope of an assertion, implicit relationships, or descriptive qualifiers, are often discarded when text is reduced to atomic triples.
This loss has practical consequences in every vertical application domain: legal compliance depends on conditions and exceptions; scientific accuracy relies on qualifiers and constraints; and multilingual retrieval requires preserving grammatical and evidential nuance.
%R THIS IS FOR RELATED WORK
%To address these issues, some work seeks to recover missing information in existing KGs \cite{refining_KGs_LLMs,KGC}, while others aim to improve extraction methods \cite{llm_finetuning_triple_extraction}, yet practical and performance limitations remain. Approaches that jointly tackle KG indexing and downstream retrieval for question‑answering tasks, such as QEDB and ATLANTIC, aim to develop more effective indexing and retrieval techniques, yet the former is limited by the explicitness of QA pairs, and the latter by its domain specificity.
%Finally, TCR-QF \cite{TCR} aims to mitigate this loss by reintroducing context or inferring missing links. Yet, these methods either rely heavily on post‑hoc inference or introduce speculative connections that may compromise precision. 

Existing efforts to repair or extract KGs \cite{refining_KGs_LLMs,KGC,llm_finetuning_triple_extraction}, enhance indexing-retrieval pipelines \cite{QEDB,atlantic}, or reintroduce missing context \cite{TCR} still suffer from limitations such as reliance on explicitly asked QA pairs, domain‑restricted applicability, and post‑hoc or speculative inference that risks reducing precision.

Rather than performing multi-hop reasoning at query time through explicit graph traversal, we shift relational composition to the indexing phase by compressing multi-hop evidence into entity-centric summaries. By aggregating contextual information across related triples during KG construction, the proposed approach transforms multi-hop question answering from a path-search problem into a structured retrieval-and-reasoning problem over compressed representations.

Building on this insight, we introduce {MDER-DR}, a KG-based QA framework that jointly addresses indexing and retrieval. 

The first component, {Map--Disambiguate--Enrich--Reduce (MDER)}, is an indexing methodology that minimizes contextual information loss during KG construction by generating natural language descriptions for triples and consolidating them into entity-level summaries. These summaries capture relational evidence spanning multiple hops, reducing the need for explicit graph traversal at inference time. 

The second component, {Decompose--Resolve (DR)}, is a retrieval mechanism that decomposes compositional user queries into resolvable triples and iteratively grounds them in the KG using the precomputed summaries.

Together, MDER and DR define an LLM-driven QA pipeline that decouples relational complexity from inference-time traversal while preserving the expressive richness of natural language. We empirically evaluate the proposed framework on multiple multi-hop QA benchmarks, including cross-lingual and domain-specific settings. Experimental results demonstrate consistent improvements over vector-based and graph-based RAG baselines, as well as increased robustness under language mismatch and sparse relational conditions.

In summary, our work makes the following contributions:
\begin{itemize}
    \item {MDER}, an indexing approach that compresses multi-hop relational context into entity-centric summaries during KG construction, reducing information loss and inference-time traversal.
    \item  {DR}, a retrieval mechanism that enables structured, iterative reasoning over summarized KG representations for compositional queries.
    \item An extensive empirical evaluation across datasets, domains, and languages, demonstrating the effectiveness and robustness of the proposed framework for multi-hop QA.
\end{itemize}

\begin{figure*}[t]
    \centering
    \includegraphics[width=1\linewidth]{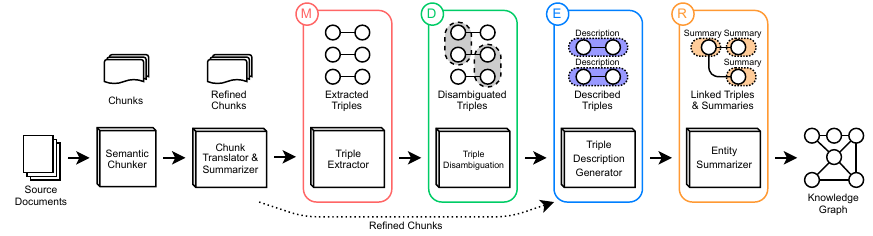}
    \caption{Indexing pipeline including the Map‑Disambiguate‑Enrich‑Reduce (MDER) workflow. It begins with preprocessing, where input documents are segmented, summarized, and translated. The core pipeline then proceeds through: (1) Map extracts subject‑predicate‑object triples; (2) Disambiguate unifies fragmented or redundant entities; (3) Enrich adds contextual descriptions to triples; and (4) Reduce generates entity-centric summaries. The final output is a structured graph of interconnected document, chunk, entity, relationship, and triple nodes, optimized for downstream KG tasks, built upon a domain-agnostic RDF/OWL ontology.}
    \label{fig:MDER}
\end{figure*}

%%%%%%%%%%%%%%%%%%%%%%%%%%
% RELATED
%%%%%%%%%%%%%%%%%%%%%%%%%%

\section{Related Work}
\label{sec:related_work}

RAG was introduced in \cite{meta_ai_first_rag} to combine the advantages of information retrieval with LLM's reasoning capabilities. It involves incorporating an information retrieval step that uses vectorized external sources such as databases, text, or web sources, depending on the specific use case, before generating responses. Microsoft took a step forward with GraphRAG \cite{microsoft_graph_rag}, which uses a KB as the external knowledge base.
Recent advances now enable LLMs to automatically construct KGs by extracting entities and relations directly from unstructured text, a process that, however, remains inherently lossy. \cite{automatic_kg_construction}. When rich language is reduced to atomic triples, key context, such as the scope of statements, implicit relationships, and important qualifiers, is often lost.

Multi-hop question answering over KGs has been extensively studied in the context of explicit path-based or iterative reasoning with methods like GraftNet~\cite{sun2018graftnet} and PullNet~\cite{sun2019pullnet}, which require explicit traversal or expansion of relational paths at inference time.
Benchmarks such as ComplexWebQuestions~\cite{talmor2018complexwebquestions} and WikiHop~\cite{welbl2018wikihop} have highlighted the challenges of decomposing questions into intermediate reasoning steps and aggregating evidence.
Neuro-symbolic approaches aim to combine the interpretability of symbolic reasoning with the flexibility of neural models, by formalizing reasoning over structured representations while learning components of the inference process~\cite{liang2017neural,dong2019neural}.  Our work is conceptually aligned with this line of research; however, instead of learning symbolic programs, we compress relational evidence into entity-centric summaries.

More recent advances in multi-hop reasoning include chain-of-thought prompting \cite{Chain}, iterative retrieval-reasoning approaches like IRCoT \cite{Interleaving} and ReAct \cite{React}, and demonstrate-search-predict frameworks \cite{Demonstrate}. While these methods excel at complex reasoning, they require multiple retrieval rounds and model calls during inference. Our approach differs by encoding multi-hop information during indexing, enabling more efficient retrieval. 

Some work seeks to recover missing information in existing KGs \cite{refining_KGs_LLMs,KGC}, while others aim to improve extraction methods \cite{llm_finetuning_triple_extraction}, yet practical and performance limitations remain. Some approaches jointly tackle KG indexing and downstream retrieval for QA tasks, such as QEDB \cite{QEDB} and ATLANTIC \cite{atlantic}, but the former is limited by the explicitness of QA pairs, and the latter by its domain specificity.
TCR-QF \cite{TCR} aims to mitigate this loss by reintroducing context or inferring missing links. However, these methods either rely heavily on post‑hoc inference or introduce speculative connections that can compromise precision. 

%Building on these lines of research, our work introduces a domain‑agnostic KG‑based QA framework that improves indexing and retrieval for multi‑hop reasoning using KG-enhanced entity-centric summaries, aiming to support robust QA across domains, languages, and data sparsity conditions.

While prior work on multi-hop and complex QA predominantly relies on explicit graph traversal, iterative subgraph expansion, or dynamic reasoning chains at query time, our approach differs by relocating relational composition to the indexing phase through entity-centric summarization, transforming multi-hop QA into a structured retrieval-and-reasoning problem over compressed representations. This perspective complements existing KGQA and knowledge-based RAG paradigms while addressing limitations related to the sparsity of graphs, scalability, and cross-lingual robustness.

%%%%%%%%%%%%%%%%%%%%%%%%%%
% METHOD
%%%%%%%%%%%%%%%%%%%%%%%%%%

\section{MDER-DR}
\label{sec:mder_dr}

We propose {Map-Disambiguate‑Enrich‑Reduce (MDER)} and {Decompose-Resolve (DR)} as two complementary paradigms building an LLM-driven, KG-based QA framework designed to enhance performance in multi-hop scenarios through MDER, a new indexing methodology that condenses relational information in entity-level summaries, and DR, a novel retrieval approach that semantically decomposes users' compositional queries into resolvable triples and grounds them into the KG using MDER’s summaries.

\subsection{Map-Disambiguate-Enrich-Reduce}
\label{sec:mder}

MDER addresses the inherent loss of contextual nuance that occurs when unstructured text is reduced to atomic triples during KG construction. Its main contribution lies in aggregating relational signals into entity‑level summaries, removing the need for downstream retrieval to traverse explicit hops. The process is structured into a preprocessing stage plus four sequential ones (i.e., mapping, disambiguation, enrichment, and reduction), each leveraging the potential of next-generation language models to incrementally refine the representation of information.
The entire approach is inspired by the \textit{divide et impera} paradigm, which, rather than relying on a few large inferences, utilises numerous small steps (e.g., language model calls), enhancing control and transparency while ensuring the consistent application of safety checks.
\Cref{fig:MDER} shows the entire indexing pipeline.

\paragraph{Preprocessing}
Before the MDER pipeline begins, the input text undergoes preprocessing, including chunking, summarization, and translation, all of which are performed using language models for text generation and embedding. This ensures that the text is structured, concise, and unified, improving consistency in the subsequent stages.

The chunking step employs a windowed semantic chunker~\cite{semantic_chunker}. The text is first split into sentences, and a sliding window then measures vector similarity between adjacent sentences. When similarity drops below a threshold (e.g., 0.75), the preceding sentences are grouped into a new chunk, and the process continues.

Chunk summarization uses a windowed mechanism that slides over the newly created chunks, summarizing the central chunk while drawing context from its neighbors. Its primary role is to prevent important passages from being fragmented, but it also produces concise, coherent representations by preserving essential meaning and removing markdown, newlines, and other extraneous noise.

Finally, translating each chunk in English provides a unified linguistic foundation for multilingual sources, ensuring consistent operation across languages. The language choice is made given that English is increasingly becoming the world's \textit{lingua franca}.

\paragraph{Map}
The pipeline begins by passing each preprocessed chunk through a prompt‑guided language model that extracts subject-predicate-object triples. Conceptually, this step maps unstructured natural language, rich in nuance and variability, into a set of atomic assertions.
Because this is the stage where information loss is most likely, the prompt is designed to encourage the detection of compound or nested relations, as well as temporal, causal, hierarchical, and synonymous links (see \Cref{sec:prompts}). This yields a more fine‑grained representation, increasing the number of triples but making implicit meaning explicit.
Extracted entities are type‑inferred from context, enabling a domain‑agnostic approach suitable for diverse applications. 

%While the prompt could also be designed to adhere to domain-specific ontologies, this approach is deliberately not pursued in this work.
%The resulting structured representation is ideal for the following pipeline steps, as it enables tracking of relationships between the original unstructured input and its output triples.

\paragraph{Disambiguate}

Following mapping, the disambiguation stage addresses redundancy and fragmentation in triple representation at the entity level. Its principal aim is to assess whether entities with syntactic or semantic differences represent, in fact, the same real-world concepts. Indeed, it's not unusual to find the same concept represented in various forms, such as synonyms, abbreviations, and grammatical variations. For example, \enquote{\textit{European Union}} and \enquote{\textit{EU}} are different syntactic forms referring to the same concept, while \enquote{\textit{Frédéric Chopin}} and \enquote{\textit{Chopin the pianist}} represent semantically distinct references to the same person. In such cases, selecting a new entity name that better aligns with the actual concept and then attributing it to all the relevant entities effectively solves the issue.

The pipeline first compares embedding similarities between candidate entity pairs. When similarity exceeds a threshold (e.g., 0.95), a prompt‑guided language model examines their original textual contexts to make the final determination (see \Cref{sec:prompts}). If the entities are deemed equivalent, it selects or generates a canonical identifier, favoring singular, concise, and general forms. These updated identifiers are then propagated back into the triples, replacing fragmented nodes with unified ones. The process repeats until no further merges occur, yielding a stable and coherent entity set.

\paragraph{Enrich}

The enrichment stage is central to MDER, reintroducing the expressive richness of natural language without compromising the graph's structural clarity and precision. Indeed, each triple is augmented with a natural‑language description that captures the contextual elements in the originating text, such as scope, qualifiers, conditions, exceptions, or temporal boundaries. To achieve this, the system retrieves the specific chunk(s) in which the triple was generated, along with the preceding and following ones, preserving local context and maintaining dependencies that span multiple chunks. Then, a prompt-guided language model produces a relevant English description that helps characterize each triple (see \Cref{sec:prompts}). 

As an example, consider a triple such as \enquote{\textit{Ferrero INTRODUCED Nutella}}. Suppose one chunk states that \enquote{\textit{Nutella was launched in 1964 as a new spread developed from earlier hazelnut products}}, while another chunk independently notes that \enquote{\textit{although Nutella was introduced in 1964, the recipe was modified in 2015 to improve spreadability}}. The enrichment step would then yield a description such as: \enquote{\textit{Ferrero introduced Nutella in 1964, and the recipe was subsequently adjusted in 2015 to enhance its texture}}.

% This step prepares the ground for the final reduction stage, which condenses relational information into compact entity‑level summaries and allows for subsequent hop-free retrieval. 

\paragraph{Reduce}

The final stage in MDER retains the contextual depth introduced during enrichment by consolidating the triple descriptions into condensed, entity-centric summaries, thereby \textit{de facto} transforming the KG into a structure that supports hop‑free downstream retrieval even for complex compositional queries.

For each entity, the system collects all triples in which it appears and concatenates their natural‑language descriptions into a single contextual element, further augmenting it by the entity’s inferred type.
A language model then generates an English summary that characterizes the entity based on this aggregated information (see \Cref{sec:prompts}).

As an example, suppose the target entity is \enquote{\textit{Marconi}}, typed as an \enquote{\textit{Engineer}}, with composing triples such as \enquote{\textit{Marconi ACHIEVED Transatlantic Wireless Signal}} and \enquote{\textit{Marconi WON Nobel Prize}} whose respective descriptions state \enquote{\textit{Marconi achieved the first transatlantic wireless signal in 1901}}, and \enquote{\textit{Marconi WON Nobel Prize in Physics in 1909}}. A language model then produces a concise summary, such as \enquote{\textit{Marconi was an engineer who achieved the first transatlantic radio transmission in 1901 and received the 1909 Nobel Prize in Physics}}.

\subsection{KG Creation}

The following sections describe the KG creation process, including the ontology definition and the KG population.

\paragraph{Ontology Definition}

The KG is built on a domain-agnostic RDF/OWL ontology that enables the representation of documents, their chunks, extracted entities, relationships, and corresponding triples.
It defines five core classes that comprise the conceptual model (i.e., \texttt{Document}, \texttt{Chunk}, \texttt{Entity}, \texttt{Relationship}, and \texttt{Triple}), as well as the object properties that connect instances of those classes and the data type properties that store literal information, such as names, URIs, and descriptions.

At the document level, documents are linked to their constituent chunks through the \texttt{hasChunk} relation, while their entities and relationships are kept using \texttt{hasEntity} and \texttt{hasRelationship}. An \texttt{xsd:string} references the \texttt{hasUri}, which may be either a unique identifier or a URL if the document is a web page, while \texttt{belongsToDocument} links chunks, entities, and relationships to their originating document.

At the chunk level, \texttt{hasNext} and \texttt{hasPrevious} preserve the original ordering, while chunk textual contents are stored in an \texttt{xsd:string}, \texttt{hasContent}. \texttt{belongsToChunk} anchors entities, relationships, and triples to their source chunk.

At the triple level, statements are represented in the graph via reification \cite{reification}, enabling metadata attachment and ensuring unique identification. Each triple references its constituent elements with \texttt{hasSource}, \texttt{hasRelationship}, and \texttt{hasTarget}. Since one of the key ideas of MDER is to enrich triples with their own descriptions, these are stored using the \texttt{xsd:string} \texttt{hasDescription}. %\texttt{composes} links entities and relationships to their originating triples.

Entities and relationships are assigned a \texttt{xsd:string} value via the \texttt{hasName} attribute, storing their unescaped name. The \texttt{xsd:string} \texttt{hasSummary} stores the entity-centric summaries. \texttt{relatesSource} and \texttt{relatesTarget} describe how entities are connected one to another, while \texttt{hasSource}/\texttt{isSourceOf} and \texttt{hasTarget}/\texttt{isTargetOf} attach entities to relationships and vice-versa.

\paragraph{KG Population}

After defining and loading the ontology, the system iterates over each unique source document, creating a corresponding \texttt{Document} node, assigning it a sanitized identifier, and storing its URI.
The system then processes all chunks belonging to that document, each of which becomes a \texttt{Chunk} node with its own URI and textual content.
Within each chunk, the system processes the assertions, generating two \texttt{Entity} nodes, one for the source (representing the subject) and one for the target (representing the object), along with a \texttt{Relationship} node to represent the predicate. These are uniquely identified by URIs derived from their names, which are titled with the unescaped form of those names. 
The system materializes each assertion as a \texttt{Triple} node, uniquely identified by a URI that concatenates its source entity, relationship, and target entity, and enriched with a description from its originating chunk. A generated summary is then generated for each entity using its triple descriptions.

\begin{figure}[t]
    \centering
    \includegraphics[width=1\linewidth]{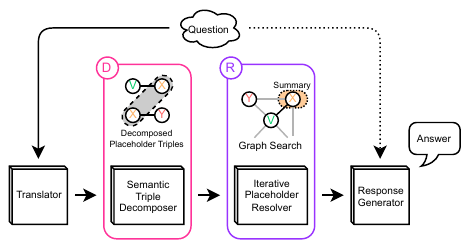}
    \caption{Retrieval pipeline, including the Decompose-Resolve (DR) approach. It begins with a user query, which is translated into English. DR then produces a set of placeholder-augmented triples from that question and grounds each placeholder in KG entities through a retrieval and reasoning loop. Resolved entities are propagated across triples, and supporting evidence is accumulated throughout the resolution process. Once all placeholders are resolved, a final answer is synthesized using accumulated summaries.}
    \label{fig:DR}
\end{figure}

\subsection{Decompose-Resolve}

Decompose‑Resolve (DR) enables the downstream QA retrieval task to extract evidence through a structured process that interprets user queries, decomposes them into semantically meaningful components, and iteratively resolves them against the KG until contextually grounded answers emerge.
The resolved entities, along with their MDER-generated summaries, are then fed into the final module that generates the user’s answer.
In this way, relational information is integrated without requiring explicit multi-hop traversals, enabling compositional queries to be resolved through direct lookups rather than graph‑walking.
\Cref{fig:DR} shows the retrieval pipeline.

\paragraph{Preprocessing}

The system first applies a language‑normalization preprocessing step that translates the user’s query into English, if not, mirroring the MDER approach and ensuring the system operates on a consistent linguistic substrate.

\paragraph{Decompose}

A %first \textit{pars destruens},
decomposition phase, more precisely Semantic Triple Decomposition, transforms the natural‑language query from the user into a set of structured triples that may contain placeholders representing unknown or intermediate entities.
This is performed through a prompt that enables the identification of the latent relational structure implied by the query (see \Cref{sec:prompts}).
Similar to the MDER mapping stage, it extracts triples from the user's intent rather than from text chunks.
%The placeholders will serve as navigational anchors to reason over multi-hop relationships without requiring the user to explicitly specify intermediate steps.

For example, a query such as \enquote{\textit{Who was the wife of the king of Ithaca?}} may be decomposed into the placeholder‑augmented triples \enquote{\textit{Ithaca HAS\_KING X}} and \enquote{\textit{X HAS\_WIFE Y}}.

\paragraph{Resolve}

The subsequent concept resolution phase, called Iterative Placeholder Resolution, progressively replaces placeholders by grounding them in KG entities.
Triples are processed sequentially in the order defined during decomposition. For each triple, the known entity serves as the retrieval anchor, and candidate KG entities are identified through an embedding‑similarity search over entity names.  For example, resolving \enquote{\textit{X}} in \enquote{\textit{Ithaca HAS\_KING X}} amounts to retrieving entities associated with \enquote{\textit{Odysseus}}.

Candidates are ranked by similarity, and the top‑$k$ entities, along with their MDER‑generated entity-level summaries, are retained. Since these summaries already capture multi‑hop information, no graph traversal is required.
A language model then analyzes and reasons over these summaries in the context of the surrounding triples, using them to resolve the current placeholder (see \Cref{sec:prompts}). Once a placeholder is resolved, its value is propagated across all triples.
The process repeats until all placeholders are consumed. Throughout this loop, summaries supporting each resolution are accumulated and capped at a maximum (e.g., 30), ensuring the final output remains reproducible.

\paragraph{Response Generation}

Once the resolution loop concludes, the system enriches the retrieved evidence by adding more summaries obtained via a similarity search between the user’s question embedding and the MDER entity‑level summaries. This optional step, which typically contributes up to 10 summaries, improves robustness to noise and helps recover relevant information even when decomposition fails to produce valid triples, as in single‑word queries.

All accumulated summaries are then passed to a final prompt‑guided component, which receives both the user’s question and the curated set of summaries and synthesizes them into the final answer for the user (see \Cref{sec:prompts}).
The generation process avoids the injection of knowledge from the model and over-interpretation.

%%%%%%%%%%%%%%%%%%%%%%%%%%
% EXPERIMENTS
%%%%%%%%%%%%%%%%%%%%%%%%%%

\section{Experiments}
\label{sec:experiments}

To assess the performance of the proposed MDER-DR architecture in multi-hop QA tasks under cross-domain shifts and across multiple languages, we compare it against common established QA approaches, including standard graph- and vector-based RAG implementations.

\subsection{Datasets}

We select multiple established multi-hop QA datasets, each with a distinct domain and reasoning style, and randomly sample 500 QA pairs from each. The associated documents are about 5,000 per dataset. {WikiQA} \cite{wikiqa} contains factual questions derived from Wikipedia and is widely used to evaluate retrieval and short answer generation;
{HotpotQA} \cite{hotpotqa} requires reasoning on multiple paragraphs with explicit evidence aggregation, making it a standard benchmark for multi-hop QA.
%{ConcurrentQA} \cite{concurrentqa} contains questions that require resolving temporal or contextual conflicts across documents, testing the ability to integrate inconsistent or evolving information.

To evaluate performance in a specialized domain, we test BenchEE, the domain-specific energy-related dataset introduced by \cite{campi_icwe}, comprising 101 pairs and their documents, which involve expert-level factual content and require technical knowledge. To this end, in addition to the automatic metrics described below, we include a comparison with a human-based evaluation conducted by domain experts.

All experiments are repeated across multiple languages (e.g., English, Italian, French, and Spanish), distinguishing between cases where the query language matches the indexing and retrieval language and cases where it does not. QA pairs are translated using a prompt-guided language model (see \Cref{sec:prompts}) and manually quality-checked. In scenarios involving mismatched languages, all mismatching languages are selected and their results averaged.

\subsection{Models}

For all tasks involving language model generation, including indexing, retrieval, and evaluation‑related outputs, we employ the \texttt{gpt‑oss:120b} or \texttt{gpt‑oss:20b} models, chosen based on task complexity, which provide the strongest performance among the open‑source LLMs \cite{gptoss}. For all embedding‑based components, we rely on the \texttt{mxbai‑embed‑large} model, selected for its state‑of‑the‑art retrieval quality and robustness across languages \cite{mxbai}.

\subsection{Compared Architectures}

We compare MDER-DR against representative baselines that capture the dominant paradigms in retrieval‑augmented QA.
First, we implement a standard {vector-based RAG} using a fixed‑size chunker (chunk size 1000 tokens, 200‑token overlap) and a top-10 chunks retriever.
Then, we include a {graph‑based} retrieval architecture based on GraphRAG \cite{microsoft_graph_rag}, using their default parametrization. Finally, we compare the results with an LLM-only, non-RAG tool (ChatGPT).

\begin{table*}[h]
    \centering
    \caption{Evaluation scores on WikiQA, HotpotQA, and BenchEE grouped by evaluation method and language match/mismatch. $\Delta (\%)$ indicates the percentage improvement in the score of MDER-DR compared with the best baseline.}
    \label{tab:multi_hop_bench}
    \begin{tabular}{llcccccc}
        \toprule
        \textbf{Dataset} & \textbf{Evaluator} & \textbf{Match} & \textbf{LLM-only} & \textbf{Vector-RAG} & \textbf{GraphRAG} & \textbf{MDER-DR} & \textbf{$\Delta$ (\%)}\\
        \midrule
        WikiQA & LLM-as-a-Judge & $\checkmark$ & 0.412 $\pm$ 0.021 & 0.477 $\pm$ 0.022 & 0.422 $\pm$ 0.022 & \underline{0.790 $\pm$ 0.012} & 66\%\\
        WikiQA & LLM-as-a-Judge &  & 0.363 $\pm$ 0.012 & 0.476 $\pm$ 0.013 & 0.424 $\pm$ 0.013 & \underline{0.698 $\pm$ 0.011} & 47\%\\
        WikiQA & Soft EM & $\checkmark$ & 0.510 $\pm$ 0.022 & 0.538 $\pm$ 0.022 & 0.510 $\pm$ 0.022 & \underline{0.800 $\pm$ 0.018} & 49\%\\
        WikiQA & Soft EM &  & 0.408 $\pm$ 0.013 & 0.538 $\pm$ 0.013 & 0.516 $\pm$ 0.013 & \underline{0.699 $\pm$ 0.012} & 30\%\\
        \midrule
        HotpotQA & LLM-as-a-Judge & $\checkmark$ & 0.565 $\pm$ 0.021 & 0.610 $\pm$ 0.021 & 0.606 $\pm$ 0.021 & \underline{0.667 $\pm$ 0.021} & 9\% \\
        HotpotQA & LLM-as-a-Judge &  & 0.515 $\pm$ 0.012 & 0.578 $\pm$ 0.013 & 0.582 $\pm$ 0.012 & \underline{0.659 $\pm$ 0.012} & 13\%\\
        HotpotQA & Soft EM & $\checkmark$ & 0.458 $\pm$ 0.022 & 0.548 $\pm$ 0.022 & 0.562 $\pm$ 0.022 & \underline{0.586 $\pm$ 0.022} & 4\% \\
        HotpotQA & Soft EM &  & 0.347 $\pm$ 0.012 & 0.471 $\pm$ 0.013 & 0.473 $\pm$ 0.013 & \underline{0.525 $\pm$ 0.013} & 11\% \\
        \midrule
        BenchEE & LLM-as-a-Judge & $\checkmark$ & 0.748 $\pm$ 0.032 & 0.768 $\pm$ 0.036 & 0.730 $\pm$ 0.035 & \underline{0.829 $\pm$ 0.031} & 8\% \\
        BenchEE & LLM-as-a-Judge &  & 0.727 $\pm$ 0.031 & 0.630 $\pm$ 0.037 & 0.681 $\pm$ 0.034 & \underline{0.830 $\pm$ 0.030} & 14\% \\
        \bottomrule
    \end{tabular}
\end{table*}

\subsection{Evaluation Methods}

The evaluation assesses answer quality using three complementary methods, each producing scores in the $[0,1]$ range.
{LLM-as-a-Judge} evaluation, similar to \cite{llms_judge}, requires a language model to assess the question, the expected answer, and the generated response, and to assign a score based on relevance and faithfulness. It adapts the prompt to the language of each instance and includes explicit rules for penalizing missing, off‑topic, contradictory, or hallucinated content (see \Cref{sec:prompts});
{Soft Exact Match (Soft EM)}, scores 1 if the normalized expected answer (i.e., after lowercasing, Unicode normalization, punctuation removal, article removal, and whitespace collapsing) appears \textit{within} the generated response, 0 otherwise.
For domain-specific QA datasets, a {human-based evaluation} employs domain experts to assign F1 scores, computed by separately assessing precision and recall between the facts in the expected answer and those in the generated response, for capturing correctness and information preservation, similar to \cite{campi_icwe}.

We utilized the bootstrap method to estimate the sampling distribution of the results. For each combination of dataset, architecture, and language, we generated $k=10000$ resampled datasets by sampling with replacement from the original test scores. We report the mean of these bootstrap samples as the estimated performance and the standard deviation as the measure of uncertainty (standard error).

\subsection{Hardware and Runtime}

All computations were performed on an NVIDIA RTX PRO 6000 Blackwell GPU (96~GB VRAM), a 48-core  Intel(R) Xeon(R) w7-2495X processor, and 125~GB of system memory. Each benchmark dataset required $\approx 1.5$ days to train using MDER-DR, while inference typically took $\approx 5$ seconds to produce an answer on our hardware. 

While MDER-DR introduces additional upfront cost during indexing due to multiple LLM calls, this cost is amortized across queries. At inference time, retrieval avoids explicit graph traversal, resulting in predictable and bounded latency.

%%%%%%%%%%%%%%%%%%%%%%%%%%
% RESULTS
%%%%%%%%%%%%%%%%%%%%%%%%%%

\section{Results}

In this section, we analyze the performance of MDER-DR relative to other RAG and non-RAG baselines, examining the impact of language and domain complexity.

%Across all the datasets, MDER‑DR consistently outperforms all baseline architectures in both LLM‑as‑a‑judge and Soft‑EM evaluations. In WikiQA, where questions typically require shorter factual answers, results highlight the limitations of traditional architectures. In HotpotQA, all RAG systems achieve relatively similar scores, yet MDER‑DR maintains a clear margin over vector‑based and graph‑based RAG approaches. The LLM-only baseline achieves lower results in both benchmarks. These trends hold consistently across all tested languages, including mismatched query–retrieval scenarios, demonstrating MDER-DR's strong cross‑lingual generalization. The results are shown in \Cref{tab:multi_hop_bench}.

Table~\ref{tab:multi_hop_bench} presents the quantitative results. MDER-DR consistently outperforms all baselines. The performance gap between MDER-DR and alternative approaches is higher in WikiQA when using the Soft EM evaluation (0.800 vs. 0.538 for Vector-RAG, the best baseline architecture). This result suggests that MDER-DR's entity-centric summaries effectively preserve the details needed for exact answer extraction. In HotpotQA, where all RAG architectures have similar performances, MDER-DR still maintains an advantage. The language mismatch results reveal MDER-DR's superior cross-lingual generalization: while other methods can experience significant performance drops when languages differ, MDER-DR exhibits smaller degradation (e.g., from 0.790 to 0.698 in WikiQA when evaluated with LLM-as-a-Judge), which can be attributed to its translation-based preprocessing and language-agnostic KG representation. Finally, the domain-specific BenchEE results confirm that MDER-DR's advantages extend to specialized contexts, achieving scores of $\approx 0.83$ in both language conditions, while GraphRAG and Vector-RAG are more sensitive to language mismatch. The consistency of MDER-DR's superior performance across diverse datasets, metrics, and linguistic conditions validates its effectiveness as a general-purpose multi-hop QA framework. Paired bootstrapping (with $k=10000$ resampled datasets and 1$\sigma$ significance threshold) reveals that the improvement introduced by MDER-DR is statistically significant in the majority of the experiments, the only exception being BenchEE evaluation with LLM-as-a-Judge using a matching language, where it is statistically indistinguishable from the LLM-only and Vector-RAG cases.

Overall, MDER-DR has better median scores compared to the alternative architectures, especially on the WikiQA dataset. Moreover, it has similar standard deviations compared with other architectures,  indicating reliability across different datasets and configurations. \Cref{fig:boxplots} (Section \ref{sec:box}) presents the performance variations across architectures in more detail.

\paragraph{Language comparison}
We evaluated whether language mismatch between queries and retrieved documents affects retrieval performance. As shown in \Cref{tab:language}, both evaluation metrics show minimal performance differences. For Soft EM, language match led to $0.564 \pm 0.103$, compared to $0.497 \pm 0.104$ for mismatch, while LLM-as-a-Judge scores were $0.635 \pm 0.144$ and $0.597 \pm 0.135$ respectively. The similar performance across conditions, coupled with overlapping confidence intervals, suggests that language mismatch does not introduce statistically significant degradation in retrieval quality.

\begin{table}[htbp]
\centering
\caption{Performance comparison based on language match vs. mismatch according to LLM-as-a-Judge and Soft EM methods.}
\label{tab:language}
\begin{tabular}{lcc}
\toprule
\textbf{Metric} & \textbf{Match} & \textbf{Mismatch} \\
\midrule
Soft EM & $0.564 \pm 0.103$ & $0.497 \pm 0.104$ \\
LLM-as-a-Judge & $0.635 \pm 0.144$ & $0.597 \pm 0.135$ \\
\bottomrule
\end{tabular}
\end{table}

%\paragraph{Comparison vs. commercial and non-RAG systems}
%As shown in Tables \ref{tab:human_benchee} and \ref{tab:judge_benchee}, the MDER architecture demonstrates competitive performance against established commercial RAG solutions. Moreover, it significantly outperforms the non-RAG baselines, which are not provided with context documents, and Graph-RAG, which indicates a classical Graph-RAG architecture. The non-RAG systems frequently suffered from hallucinations, whereas the retrieval-augmented approaches grounded their answers in the information contained in the context documents. 

\paragraph{LLM-as-a-Judge vs. human evaluation}
To validate our automated metrics, we calculated the correlation between the LLM-based judge scores and the human evaluation scores on the BenchEE dataset. We observed a strong Pearson correlation in the scores ($\approx 0.79$), which indicates that the LLM judge is a reliable proxy for human evaluations. \Cref{tab:human} shows the human evaluation scores on the BenchEE.

\begin{table}[htbp]
\centering
\caption{Human evaluation results on BenchEE, based on judgments from three energy‑domain experts.}
\label{tab:human}
\begin{tabular}{lcc}
\toprule
\textbf{Architecture} & \textbf{Match} & \textbf{Mismatch} \\
\midrule
LLM-only & $0.696 \pm 0.029$ & $0.636 \pm 0.029$ \\
Vector-RAG & $0.762 \pm 0.034$ & $0.624 \pm 0.038$ \\
GraphRAG & $0.775 \pm 0.036$ & $0.734 \pm 0.039$ \\
\underline{MDER-DR} & \underline{$0.910 \pm 0.023$} & \underline{$0.909 \pm 0.023$} \\
\bottomrule
\end{tabular}
\end{table}

\section{Conclusions}
The MDER-DR framework addresses the inherent information loss in conventional KG indexing by shifting relational composition to the indexing phase. The MDER approach preserves context by consolidating triple-level descriptions into entity-centric summaries. The DR mechanism enables the resolution of compositional queries via structured retrieval rather than explicit graph walking. 

Empirical results show that MDER-DR  outperforms standard RAG baselines across multiple datasets. Specifically, in WikiQA, the system achieved a score of 0.800 compared to 0.538 for Vector-RAG. Furthermore, the framework is robust to language mismatch, showing only small performance degradation, which is attributed to its translation-based preprocessing and language-agnostic representation.

Future research directions include: investigating methods to reduce the computational cost of multiple LLM calls during the MDER indexing phase; exploring the application of the framework to larger datasets; and conducting an ablation study of the architecture's components and hyperparameter tuning.

\section*{Impact Statement}
This paper presents work aimed at advancing the field of Machine Learning. There are many potential societal consequences of our work, none of which we feel must be specifically highlighted here.

\section*{Use of Generative AI Tools}
LLMs were used for proofreading and grammatical corrections of the manuscript only. They were not used to generate scientific content, experiments, analyses, or conclusions. 

%%%%%%%%%%%%%%%%%%%%%%%%%%
% RELATED
%%%%%%%%%%%%%%%%%%%%%%%%%%


\bibliography{main}

@inproceedings{sun2018graftnet,
  title     = {GraftNet: Graph-based Reasoning for Question Answering over Knowledge Graphs},
  author    = {Sun, Haitian and Ma, Tuan and Yih, Wen-tau and Tsai, Chen-Tse and Liu, Jingjing and Chang, Ming-Wei},
  booktitle = {Proceedings of the 2018 Conference on Empirical Methods in Natural Language Processing},
  pages     = {2506--2515},
  year      = {2018}
}

@inproceedings{sun2019pullnet,
  title     = {PullNet: Open Domain Question Answering with Iterative Retrieval},
  author    = {Sun, Haitian and Dhingra, Bhuwan and Chang, Ming-Wei and Cohen, William W.},
  booktitle = {Proceedings of the 2019 Conference on Empirical Methods in Natural Language Processing},
  pages     = {2380--2390},
  year      = {2019}
}

@inproceedings{talmor2018complexwebquestions,
  title     = {The Web as a Knowledge Base for Answering Complex Questions},
  author    = {Talmor, Alon and Berant, Jonathan},
  booktitle = {Proceedings of the 56th Annual Meeting of the Association for Computational Linguistics},
  pages     = {641--651},
  year      = {2018}
}

@inproceedings{welbl2018wikihop,
  title     = {Constructing Datasets for Multi-hop Reading Comprehension Across Documents},
  author    = {Welbl, Johannes and Stenetorp, Pontus and Riedel, Sebastian},
  booktitle = {Proceedings of the 2018 Conference on Empirical Methods in Natural Language Processing},
  pages     = {512--521},
  year      = {2018}
}

@inproceedings{liang2017neural,
  title     = {Neural Symbolic Machines: Learning Semantic Parsers on Freebase with Weak Supervision},
  author    = {Liang, Chen and Norouzi, Mohammad and Berant, Jonathan and Le, Quoc V. and Lao, Ni},
  booktitle = {Proceedings of the International Conference on Learning Representations},
  year      = {2017}
}

@inproceedings{dong2019neural,
  title     = {Neural Logic Machines},
  author    = {Dong, Honghua and Mao, Jiayuan and Lin, Tian and Wang, Chong and Li, Lihong and Zhou, Denny},
  booktitle = {Proceedings of the International Conference on Learning Representations},
  year      = {2019}
}

@article{KGs_aidan,
    author = {Hogan, Aidan and Blomqvist, Eva and Cochez, Michael and D’amato, Claudia and Melo, Gerard De and Gutierrez, Claudio and Kirrane, Sabrina and Gayo, Jos\'{e} Emilio Labra and Navigli, Roberto and Neumaier, Sebastian and Ngomo, Axel-Cyrille Ngonga and Polleres, Axel and Rashid, Sabbir M. and Rula, Anisa and Schmelzeisen, Lukas and Sequeda, Juan and Staab, Steffen and Zimmermann, Antoine},
    title = {{Knowledge Graphs}},
    year = {2021},
    issue_date = {May 2022},
    publisher = {Association for Computing Machinery},
    address = {New York, NY, USA},
    volume = {54},
    number = {4},
    issn = {0360-0300},
    doi = {10.1145/3447772},
    journal = {ACM Comput. Surv.},
    month = jul,
    articleno = {71},
    numpages = {37},
}

@inproceedings{Interleaving,
  title={Interleaving retrieval with chain-of-thought reasoning for knowledge-intensive multi-step questions},
  author={Trivedi, Harsh and Balasubramanian, Niranjan and Khot, Tushar and Sabharwal, Ashish},
  booktitle={Proceedings of the 61st annual meeting of the association for computational linguistics (volume 1: long papers)},
  pages={10014--10037},
  year={2023}
}

@inproceedings{React,
  title={React: Synergizing reasoning and acting in language models},
  author={Yao, Shunyu and Zhao, Jeffrey and Yu, Dian and Du, Nan and Shafran, Izhak and Narasimhan, Karthik R and Cao, Yuan},
  booktitle={The eleventh international conference on learning representations},
  year={2022}
}

@article{Demonstrate,
  title={Demonstrate-search-predict: Composing retrieval and language models for knowledge-intensive nlp},
  author={Khattab, Omar and Santhanam, Keshav and Li, Xiang Lisa and Hall, David and Liang, Percy and Potts, Christopher and Zaharia, Matei},
  journal={arXiv preprint arXiv:2212.14024},
  year={2022}
}

@inproceedings{Chain,
 author = {Wei, Jason and Wang, Xuezhi and Schuurmans, Dale and Bosma, Maarten and Ichter, Brian and Xia, Fei and Chi, Ed and Le, Quoc V and Zhou, Denny},
 booktitle = {Advances in Neural Information Processing Systems},
 editor = {S. Koyejo and S. Mohamed and A. Agarwal and D. Belgrave and K. Cho and A. Oh},
 pages = {24824--24837},
 publisher = {Curran Associates, Inc.},
 title = {Chain-of-Thought Prompting Elicits Reasoning in Large Language Models},
 volume = {35},
 year = {2022}
}

@inproceedings{TCR,
    author = {Huang, Manzong and Bu, Chenyang and He, Yi and Wu, Xindong},
    title = {{How to mitigate information loss in knowledge graphs for graphrag: leveraging triple context restoration and query-driven feedback}},
    year = {2025},
    isbn = {978-1-956792-06-5},
    doi = {10.24963/ijcai.2025/901},
    booktitle = {Proceedings of the Thirty-Fourth International Joint Conference on Artificial Intelligence},
    articleno = {901},
    numpages = {9},
    location = {Montreal, Canada},
    series = {IJCAI '25}
}

@inproceedings{KGC,
    author = {Zhao, Wei and Zhang, Ziyang},
    title = {{Knowledge Graph Completion Fusing Description and Structural Feature}},
    year = {2025},
    isbn = {9798400713231},
    publisher = {Association for Computing Machinery},
    address = {New York, NY, USA},
    doi = {10.1145/3728199.3728226},
    booktitle = {Proceedings of the 2025 3rd International Conference on Communication Networks and Machine Learning},
    pages = {178–184},
    numpages = {7},
    location = {
    },
    series = {CNML '25}
}

@Online{semantic_chunker,
  accessed = {2025-12-12},
  author   = {{Kamradt, Greg}},
  title    = {{5 Levels Of Text Splitting}},
  year = {2024},
}

@inproceedings{llm_finetuning_triple_extraction,
    title = "Fine-tuning Language Models for Triple Extraction with Data Augmentation",
    author = "Zhang, Yujia  and
      Sadler, Tyler  and
      Taesiri, Mohammad Reza  and
      Xu, Wenjie  and
      Reformat, Marek",
    editor = "Biswas, Russa  and
      Kaffee, Lucie-Aim{\'e}e  and
      Agarwal, Oshin  and
      Minervini, Pasquale  and
      Singh, Sameer  and
      de Melo, Gerard",
    booktitle = "Proceedings of the 1st Workshop on Knowledge Graphs and Large Language Models (KaLLM 2024)",
    month = aug,
    year = "2024",
    address = "Bangkok, Thailand",
    publisher = "Association for Computational Linguistics",
    doi = "10.18653/v1/2024.kallm-1.12",
    pages = "116--124",
}

@article{automatic_kg_construction,
author = {Zhong, Lingfeng and Wu, Jia and Li, Qian and Peng, Hao and Wu, Xindong},
title = {{A Comprehensive Survey on Automatic Knowledge Graph Construction}},
year = {2023},
issue_date = {April 2024},
publisher = {Association for Computing Machinery},
address = {New York, NY, USA},
volume = {56},
number = {4},
issn = {0360-0300},
doi = {10.1145/3618295},
journal = {ACM Comput. Surv.},
month = nov,
articleno = {94},
numpages = {62},
}

@ARTICLE{LLMs_and_KGs,
  author={Pan, Shirui and Luo, Linhao and Wang, Yufei and Chen, Chen and Wang, Jiapu and Wu, Xindong},
  journal={IEEE Transactions on Knowledge and Data Engineering}, 
  title={{Unifying Large Language Models and Knowledge Graphs: A Roadmap}}, 
  year={2024},
  volume={36},
  number={7},
  pages={3580-3599},
  doi={10.1109/TKDE.2024.3352100}}

@inproceedings{refining_KGs_LLMs,
    title = "Refining Noisy Knowledge Graph with Large Language Models",
    author = "Dong, Na  and
      Kertkeidkachorn, Natthawut  and
      Liu, Xin  and
      Shirai, Kiyoaki",
    editor = "Gesese, Genet Asefa  and
      Sack, Harald  and
      Paulheim, Heiko  and
      Merono-Penuela, Albert  and
      Chen, Lihu",
    booktitle = "Proceedings of the Workshop on Generative AI and Knowledge Graphs (GenAIK)",
    month = jan,
    year = "2025",
    address = "Abu Dhabi, UAE",
    publisher = "International Committee on Computational Linguistics",
    pages = "78--86"
}

@article{QEDB, title={{QA Is the New KR: Question-Answer Pairs as Knowledge Bases}}, volume={37}, DOI={10.1609/aaai.v37i13.26794}, number={13}, journal={Proceedings of the AAAI Conference on Artificial Intelligence}, author={Cohen, William W. and Chen, Wenhu and De Jong, Michiel and Gupta, Nitish and Presta, Alessandro and Verga, Pat and Wieting, John}, year={2024}, month={Jul.}, pages={15385-15392} }

@article{atlantic,
  title={{Atlantic: Structure-aware retrieval-augmented language model for interdisciplinary science}},
  author={Munikoti, Sai and Acharya, Anurag and Wagle, Sridevi and Horawalavithana, Sameera},
  journal={arXiv preprint arXiv:2311.12289},
  year={2023}
}

@article{reification,
  title={Reified Input/Output logic: Combining Input/Output logic and Reification to represent norms coming from existing legislation},
  author={Robaldo, Livio and Sun, Xin},
  journal={Journal of Logic and Computation},
  volume={27},
  number={8},
  pages={2471--2503},
  year={2017},
  publisher={Oxford University Press (OUP)}
}

@inproceedings{llms_judge,
 author = {Zheng, Lianmin and Chiang, Wei-Lin and Sheng, Ying and Zhuang, Siyuan and Wu, Zhanghao and Zhuang, Yonghao and Lin, Zi and Li, Zhuohan and Li, Dacheng and Xing, Eric and Zhang, Hao and Gonzalez, Joseph E and Stoica, Ion},
 booktitle = {Advances in Neural Information Processing Systems},
 editor = {A. Oh and T. Naumann and A. Globerson and K. Saenko and M. Hardt and S. Levine},
 pages = {46595--46623},
 publisher = {Curran Associates, Inc.},
 title = {{Judging LLM-as-a-Judge with MT-Bench and Chatbot Arena}},
 volume = {36},
 year = {2023}
}

@misc{microsoft_graph_rag,
	title        = {From Local to Global: A Graph RAG Approach to Query-Focused Summarization},
	author       = {Darren Edge and Ha Trinh and Newman Cheng and Joshua Bradley and Alex Chao and Apurva Mody and Steven Truitt and Dasha Metropolitansky and Robert Osazuwa Ness and Jonathan Larson},
	year         = 2025,
	month        = {Apr},
	url          = {https://arxiv.org/abs/2404.16130},
	eprint       = {2404.16130},
	archiveprefix = {arXiv},
	primaryclass = {cs.CL}
}

@InProceedings{campi_icwe,
author="Campi, Riccardo
and Pinciroli Vago, Nicol{\`o} Oreste
and Giudici, Mathyas
and Rodriguez-Guisado, Pablo Barrachina
and Brambilla, Marco
and Fraternali, Piero",
editor="Verma, Himanshu
and Bozzon, Alessandro
and Mauri, Andrea
and Yang, Jie",
title="A Graph-Based RAG for Energy Efficiency Question Answering",
booktitle="Web Engineering",
year="2026",
publisher="Springer Nature Switzerland",
address="Cham",
pages="41--55",
isbn="978-3-031-97207-2"
}

@inproceedings{wikiqa,
    title = "{W}iki{QA}: A Challenge Dataset for Open-Domain Question Answering",
    author = "Yang, Yi  and
      Yih, Wen-tau  and
      Meek, Christopher",
    editor = "M{\`a}rquez, Llu{\'i}s  and
      Callison-Burch, Chris  and
      Su, Jian",
    booktitle = "Proceedings of the 2015 Conference on Empirical Methods in Natural Language Processing",
    month = sep,
    year = "2015",
    address = "Lisbon, Portugal",
    publisher = "Association for Computational Linguistics",
    doi = "10.18653/v1/D15-1237",
    pages = "2013--2018"
}

@inproceedings{hotpotqa,
    title = "{H}otpot{QA}: A Dataset for Diverse, Explainable Multi-hop Question Answering",
    author = "Yang, Zhilin  and
      Qi, Peng  and
      Zhang, Saizheng  and
      Bengio, Yoshua  and
      Cohen, William  and
      Salakhutdinov, Ruslan  and
      Manning, Christopher D.",
    editor = "Riloff, Ellen  and
      Chiang, David  and
      Hockenmaier, Julia  and
      Tsujii, Jun{'}ichi",
    booktitle = "Proceedings of the 2018 Conference on Empirical Methods in Natural Language Processing",
    month = oct # "-" # nov,
    year = "2018",
    address = "Brussels, Belgium",
    publisher = "Association for Computational Linguistics",
    doi = "10.18653/v1/D18-1259",
    pages = "2369--2380",
}

@article{gptoss,
  title={{gpt-oss-120b \& gpt-oss-20b model card}},
  author={Agarwal, Sandhini and Ahmad, Lama and Ai, Jason and Altman, Sam and Applebaum, Andy and Arbus, Edwin and Arora, Rahul K and Bai, Yu and Baker, Bowen and Bao, Haiming and others},
  journal={arXiv preprint arXiv:2508.10925},
  year={2025}
}

@online{mxbai,
  title={{Open Source Strikes Bread - New Fluffy Embedding Model}},
  author={Sean Lee and Aamir Shakir and Darius Koenig and Julius Lipp},
  year={2024}
}

@article{rag_survey,
title = {A Survey on RAG with LLMs},
journal = {Procedia Computer Science},
volume = {246},
pages = {3781-3790},
year = {2024},
note = {28th International Conference on Knowledge Based and Intelligent information and Engineering Systems (KES 2024)},
issn = {1877-0509},
doi = {10.1016/j.procs.2024.09.178},
author = {Muhammad Arslan and Hussam Ghanem and Saba Munawar and Christophe Cruz}
}

@inproceedings{meta_ai_first_rag,
	title        = {Retrieval-augmented generation for knowledge-intensive NLP tasks},
	author       = {Lewis, Patrick and Perez, Ethan and Piktus, Aleksandra and Petroni, Fabio and Karpukhin, Vladimir and Goyal, Naman and K\"{u}ttler, Heinrich and Lewis, Mike and Yih, Wen-tau and Rockt\"{a}schel, Tim and Riedel, Sebastian and Kiela, Douwe},
	year         = 2020,
	booktitle    = {Proceedings of the 34th International Conference on Neural Information Processing Systems},
	location     = {Vancouver, BC, Canada},
	publisher    = {Curran Associates Inc.},
	address      = {Red Hook, NY, USA},
	series       = {NIPS '20},
	volume       = 33,
	pages        = {9459–9474},
	isbn         = 9781713829546,
	articleno    = 793,
	numpages     = 16,
	editor       = {Larochelle, H. and Ranzato, M. and Hadsell, R. and Balcan, M.F. and Lin, H.}
}
\bibliographystyle{icml2026}

%%%%%%%%%%%%%%%%%%%%%%%%%%%%%%%%%%%%%%%%%%%%%%%%%%%%%%%%%%%%%%%%%%%%%%%%%%%%%%%
%%%%%%%%%%%%%%%%%%%%%%%%%%%%%%%%%%%%%%%%%%%%%%%%%%%%%%%%%%%%%%%%%%%%%%%%%%%%%%%
% APPENDIX
%%%%%%%%%%%%%%%%%%%%%%%%%%%%%%%%%%%%%%%%%%%%%%%%%%%%%%%%%%%%%%%%%%%%%%%%%%%%%%%
%%%%%%%%%%%%%%%%%%%%%%%%%%%%%%%%%%%%%%%%%%%%%%%%%%%%%%%%%%%%%%%%%%%%%%%%%%%%%%%
\newpage
\appendix
\onecolumn

\section{Box plots}
\label{sec:box}
\begin{figure*}[!ht]
    \centering
    % Row 1: LLM-as-a-Judge, WikiQA
    \begin{subfigure}[b]{0.48\textwidth}
        \centering
        \includegraphics[width=\textwidth]{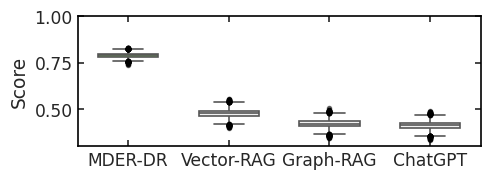}
        \caption{LLM-as-a-Judge, WikiQA (language match)}
        \label{fig:judge_wikiqa_same}
    \end{subfigure}
    \hfill
    \begin{subfigure}[b]{0.48\textwidth}
        \centering
        \includegraphics[width=\textwidth]{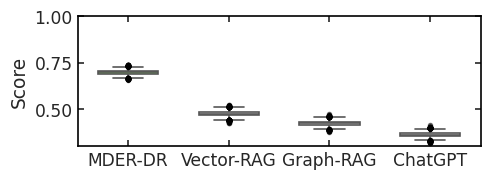}
        \caption{LLM-as-a-Judge, WikiQA (language mismatch)}
        \label{fig:judge_wikiqa_diff}
    \end{subfigure}
    
    \vspace{0.3cm}
    
    % Row 2: Exact Match, WikiQA
    \begin{subfigure}[b]{0.48\textwidth}
        \centering
        \includegraphics[width=\textwidth]{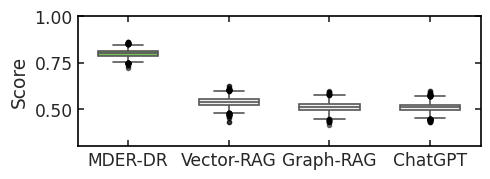}
        \caption{Soft EM, WikiQA (language match)}
        \label{fig:em_wikiqa_same}
    \end{subfigure}
    \hfill
    \begin{subfigure}[b]{0.48\textwidth}
        \centering
        \includegraphics[width=\textwidth]{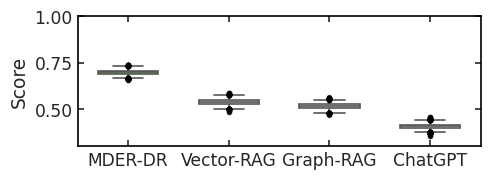}
        \caption{Soft EM, WikiQA (language mismatch)}
        \label{fig:em_wikiqa_diff}
    \end{subfigure}
    
    \vspace{0.3cm}
    
    % Row 3: LLM-as-a-Judge, HotpotQA
    \begin{subfigure}[b]{0.48\textwidth}
        \centering
        \includegraphics[width=\textwidth]{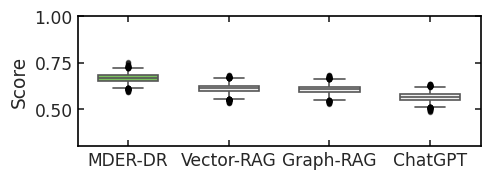}
        \caption{LLM-as-a-Judge, HotpotQA (language match)}
        \label{fig:judge_hotpot_same}
    \end{subfigure}
    \hfill
    \begin{subfigure}[b]{0.48\textwidth}
        \centering
        \includegraphics[width=\textwidth]{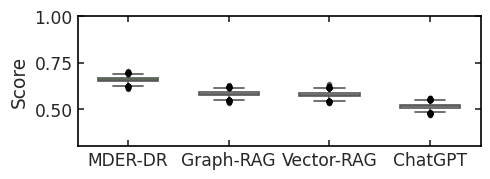}
        \caption{LLM-as-a-Judge, HotpotQA (language mismatch)}
        \label{fig:judge_hotpot_diff}
    \end{subfigure}
    
    \vspace{0.3cm}
    
    % Row 4: Soft EM, HotpotQA
    \begin{subfigure}[b]{0.48\textwidth}
        \centering
        \includegraphics[width=\textwidth]{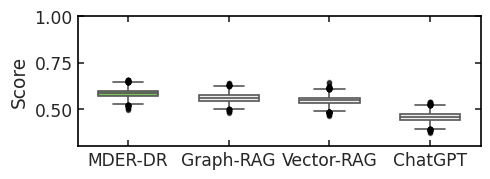}
        \caption{Soft EM, HotpotQA (language match)}
        \label{fig:em_hotpot_same}
    \end{subfigure}
    \hfill
    \begin{subfigure}[b]{0.48\textwidth}
        \centering
        \includegraphics[width=\textwidth]{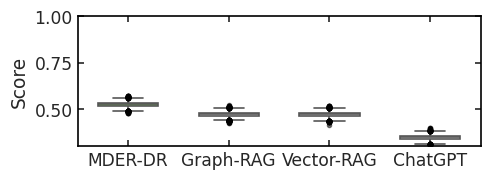}
        \caption{Soft EM, HotpotQA (language mismatch)}
        \label{fig:em_hotpot_diff}
    \end{subfigure}
    
    \vspace{0.3cm}
    
    % Row 5: LLM-as-a-Judge, BenchEE
    \begin{subfigure}[b]{0.48\textwidth}
        \centering
        \includegraphics[width=\textwidth]{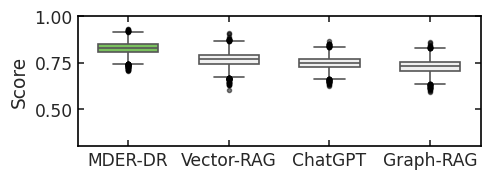}
        \caption{LLM-as-a-Judge, BenchEE (language match)}
        \label{fig:judge_benchee_same}
    \end{subfigure}
    \hfill
    \begin{subfigure}[b]{0.48\textwidth}
        \centering
        \includegraphics[width=\textwidth]{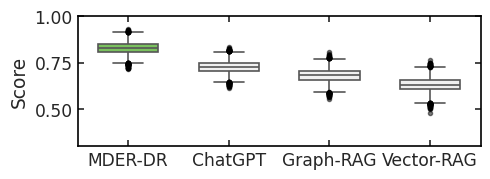}
        \caption{LLM-as-a-Judge, BenchEE (language mismatch)}
        \label{fig:judge_benchee_diff}
    \end{subfigure}
    
    \caption{Performance evaluation across different datasets and metrics. Results are shown for WikiQA, HotpotQA, and BenchEE datasets using the scores obtained with LLM-as-a-Judge and Soft EM, comparing language match vs. language mismatch conditions.}
    \label{fig:boxplots}
\end{figure*}

\clearpage
\section{Prompts}
\label{sec:prompts}

This appendix provides the full set of prompts used in our experiments.

\begin{tcolorbox}[colback=black!5!white, colframe=black!50!blue, title=Translation Prompt (Preprocessing), fontupper=\tiny]
\begin{verbatim}
---Role---
You are a system that translates the given text or question from any language into {language}.

---Goal---
Translate the given text or question into clear and accurate {language}.
DO NOT RESPOND OR ADD FABRICATED DETAILS!

---Output---
The output must not include quotation marks, formatting symbols, or any commentary.
The output must be plain text only.

---Important---
DO NOT ANSWER!
The translation should be in {language}.
If the given text is already in {language}, just return it.
\end{verbatim}
\end{tcolorbox}

\begin{tcolorbox}[colback=black!5!white, colframe=black!50!blue, title=Chunk Summarization Prompt (Preprocessing), fontupper=\tiny]
\begin{verbatim}
---Role---
You are a system that summarize the given text chunk.

---Goal---
Produce a clear, concise, and comprehensive summary of the provided text chunk.
Preserve all relevant facts, data, and context necessary for full understanding.
Eliminate redundancy and avoid unnecessary elaboration.
In case of a question, DO NOT ANSWER!

---Context---
{context}

---Output---
Output must be plain text only, without formatting symbols, commentary, or quotations.
\end{verbatim}
\end{tcolorbox}

\begin{tcolorbox}[colback=black!5!white, colframe=black!50!blue, title=Triple Extraction Prompt (Map), fontupper=\tiny]
\begin{verbatim}
# Knowledge Graph Instructions
## 1. Overview
You are a top-tier algorithm designed for extracting information in structured formats to build a knowledge graph.
Try to capture as much information from the text as possible without sacrificing accuracy.
Do not add any information that is not explicitly mentioned in the text.
- **Nodes** represent entities and concepts.
- The aim is to achieve simplicity and clarity in the knowledge graph, making it accessible for a vast audience.
## 2. Labeling Nodes
- **Consistency**: Ensure you use available types for node labels.
Ensure you use basic or elementary types for node labels.
- For example, when you identify an entity representing a person, always label it as **'person'**.
Avoid using more specific terms like 'mathematician' or 'scientist'.
- **Node IDs**: Never utilize integers as node IDs. Node IDs should be names or human-readable identifiers found in the text.
- **Relationships** represent connections between entities or concepts.
Ensure consistency and generality in relationship types when constructing knowledge graphs. 
Instead of using specific and momentary types such as 'BECAME_PROFESSOR', use more general and timeless relationship types
like 'PROFESSOR'. Make sure to use general and timeless relationship types!
## 3. Coreference Resolution
- **Maintain Entity Consistency**: When extracting entities, it's vital to ensure consistency.
If an entity, such as "John Doe", is mentioned multiple times in the text but is referred to by different names or pronouns
(e.g., "Joe", "he"), always use the most complete identifier for that entity throughout the knowledge graph.
In this example, use "John Doe" as the entity ID. Remember, the knowledge graph should be coherent and easily understandable,
so maintaining consistency in entity references is crucial.
## 4. Strict Compliance
Adhere to the rules strictly. Non-compliance will result in termination.

Ensure that:
- No detail is omitted, even if implicit or inferred
- Compound or nested relationships are captured
- Temporal, causal, or hierarchical links are included (e.g., Crime And Punishment (1998 Film) produces Crime And Punishment
HAS_YEAR 1988)
- Synonyms or aliases are noted if present (e.g., Cornelia "Lia" Melis produces Cornelia Melis HAS_ALIAS Lia)
- Types are noted and correctly separed from the entity names (e.g., Crime And Punishment (Film) produces Crime And Punishment
as entity name and Film as type)
- Prefer short and concise node and relationship names instead of keeping names (e.g., Crime and Punishment (Spanish: Crimen y
castigo) produces Crime and Punishment HAS_SPANISH_TITLE Crimen y castigo)
- Do not merge multi-word entities into single tokens (e.g., prefer Class A instead of ClassA)

Tip: Make sure to answer in the correct format and do not include any explanations.
Use the given format to extract information from the following input: {input}
\end{verbatim}
\end{tcolorbox}

\begin{tcolorbox}[colback=black!5!white, colframe=black!50!blue, title=Entity Comparator Prompt (Disambiguation), fontupper=\tiny]
\begin{verbatim}
---Role---
You are a system that determines whether two given entities represent the same concept or item.

---Goal---
Return "SAME" if the entities refer to the same concept, even if they differ in grammatical number, formatting, or minor
variations (e.g., "Deduction" vs "Deductions", "Microwave" vs "Microwaves").
Return "DIFFERENT" if the entities differ in meaning, quantity, time span, or any other substantive attribute (e.g., "6H day"
vs "8H day", "1 March 31 December" vs "1 March 15 December").
Focus on core meaning, not surface form.

---Context---
{context}

---Output---
Plain text only: either "SAME" or "DIFFERENT" with no formatting, commentary, or explanation.
\end{verbatim}
\end{tcolorbox}

\begin{tcolorbox}[colback=black!5!white, colframe=black!50!blue, title=New Entity Name Selector Prompt (Disambiguation), fontupper=\tiny]
\begin{verbatim}
---Role---
You are a system that selects the most representative name between two given entities.

---Goal---
Return the entity name that best represents the shared concept between the two inputs.
If both refer to the same concept, choose the version that is:
- Singular (not plural)
- Shorter in length
- More general or canonical
If the entities differ in meaning, return the one that is more representative or commonly used.
If neither entity name is sufficiently representative, you may select a different name that is not among the two inputs, as
long as it best captures the shared concept and is expressed in English.

---Context---
{context}

---Output---
Plain text only: the selected entity name with no formatting, commentary, or explanation.
\end{verbatim}
\end{tcolorbox}

\begin{tcolorbox}[colback=black!5!white, colframe=black!50!blue, title=Triple Description Prompt (Enrich), fontupper=\tiny]
\begin{verbatim}
---Role---
You are a system that generates concise descriptions of RDF triples using the context provided in a given text chunks.

---Goal---
Generate concise and informative descriptions for the given triple based on the provided text chunks.
The descriptions should not include any information that is not present in the provided text chunks.
The descriptions should include all relevant information from the chunks that help characterize the triple.
The descriptions should be in English.
The descriptions should be in the format "description" without quotes.

---Chunks---
{chunks}
\end{verbatim}
\end{tcolorbox}

\begin{tcolorbox}[colback=black!5!white, colframe=black!50!blue, title=Entity Summarization Prompt (Reduce), fontupper=\tiny]
\begin{verbatim}
---Role---
You are a system that generates concise descriptions of RDF entities using the context provided in a given text chunks.

---Goal---
Generate concise and informative descriptions for the given entity based on the provided text chunks.
The descriptions should not include any information that is not present in the provided text chunks.
The descriptions should include all relevant information from the chunks that helps characterize the entity.
The descriptions should be in English.
The descriptions should be in the format "description" without quotes.

---Chunks---
{chunks}
\end{verbatim}
\end{tcolorbox}

\begin{tcolorbox}[colback=black!5!white, colframe=black!50!green, title=Semantic Triple Decomposition Prompt (Decompose), fontupper=\tiny]
\begin{verbatim}
---Role---
You are a linguistic analysis system.

---Goal---
Your task is to receive a user’s question and extract information in structured formats to query a knowledge graph:
1. head: the entities or concepts the question is about.  
2. relationship: the actions, states, or relations expressed in the question.  
3. tail: the targets or complements of the relationships. It will then be retrieved in the knowledge graph.

For each question:
- Write separate sentences that clearly state the head(s), relationship(s), and tail(s) needed to answer.  
- Do not answer the question. Only decompose it into its logical components.
- Use consistent, knowledge-graph-friendly phrasing (no extra words like "the head is…").
- Use simple, explicit {language} language.
- If multiple questions are asked, produce one sentence per question.
- Use `X`, `Y`, `Z` as the placeholders for unknown tails.
- Support multi‑hop relationships: When the question refers to a property of an entity that is itself obtained through another
relation, produce multiple sentences, one per hop. Each hop must be expressed as its own knowledge‑graph triple.
- DO NOT RESPOND OR ADD FABRICATED DETAILS!

---Important---
DO NOT ANSWER!

---Examples---

Input Questions:
"Who wrote Hamlet? What is the capital of France?"

Output Sentences:
(Hamlet, WROTE_BY, X)
(France, HAS_CAPITAL, Y)

---

Input Questions:
"Who painted the Mona Lisa? Where is Mount Everest located? What language is spoken in Brazil?"

Output Sentences:
(Mona Lisa, PAINTED_BY, X)
(Mount Everest, LOCATED_IN, Y)
(Brazil, SPEAKS_LANGUAGE, Z)

---

Input Questions:
"Who died later, Hammond Innes or Adrian Solomons?"

Output Sentences:
(Hammond Innes, DIED_IN, X)
(Adrian Solomons, DIED_IN, Y)

---Multi-Hop Examples---

Input:  
"What is the place of birth of the director of film Radio Stars On Parade?"

Output:  
(Radio Stars On Parade, HAS_DIRECTOR, X)
(X, BORN_IN, Y)

---

Input:  
"What is the nationality of the author of The Name of the Rose?"

Output:
(The Name of the Rose, HAS_AUTHOR, X)
(X, HAS_NATIONALITY, Y)
\end{verbatim}
\end{tcolorbox}

\begin{tcolorbox}[colback=black!5!white, colframe=black!50!green, title=Iterative Placeholder Resolution Prompt (Resolve), fontupper=\tiny]
\begin{verbatim}
---Role---
You are an entity-relation reasoning assistant. 
You work with triples in the form (Subject, Relation, Object).

---Task---
- You will be given a list of triples and a context containing KG results.
- You have to check if the triples can be updated using the context.
- If a triple contains a placeholder (e.g., "X", "Y", "Z"), replace it with the actual entity name from the context.
- Always output the updated current triple in the format: (Subject, Relation, Object).
- Do not invent entities; only use those explicitly given in the context.
- DO NOT ADD FABRICATED DETAILS!

---Examples---

Input triples:
(In Our Lifetime, PERFORMED_BY, X)
(X, FATHER, Y)

Context:
In Our Lifetime | In Our Lifetime is a song by Marvin Gaye...

Output triple:
(In Our Lifetime, PERFORMED_BY, Marvin Gaye)
(Marvin Gaye, FATHER, Y)

---

Input triples:
(The Hobbit, WRITTEN_BY, X)
(X, BORN_IN, Y)

Context:
The Hobbit | The Hobbit is a fantasy novel written by J. R. R. Tolkien...

Output triple:
(The Hobbit, WRITTEN_BY, J. R. R. Tolkien)
(J. R. R. Tolkien, BORN_IN, Y)

---

Input triples:
(Starry Night, CREATED_BY, X)
(X, SIBLING, Y)

Context:
Starry Night | Starry Night is a painting by Vincent van Gogh...

Output triple:
(Starry Night, CREATED_BY, Vincent van Gogh)
(Vincent van Gogh, SIBLING, Y)

---Context---
{context}
\end{verbatim}
\end{tcolorbox}

\begin{tcolorbox}[colback=black!5!white, colframe=black!50!green, title=Response Generation Prompt, fontupper=\tiny]
\begin{verbatim}
---Role---
You are a helpful assistant responding to questions about data in the tables provided.
Answer in {language} 

---Goal---
Generate a response in {language} that responds to the user's question, summarizing all information in the input data tables.
If you don't know the answer, just say so. Do not make anything up.
Points supported by data should be based on the provided input data tables as follows: "This is an example sentence supported
by multiple documents."

For example:
"Person X is the owner of Company Y and subject to many allegations of wrongdoing."
Do not include information where the supporting evidence for it is not provided.

---Data tables---
{context_data}

---Target response length and format---
Answer in {language} in with a {answer_length} format of the text.
Answer to the question ONLY.
\end{verbatim}
\end{tcolorbox}

\begin{tcolorbox}[colback=black!5!white, colframe=black!50!red, title=Retrieval Failed Response Generation Prompt, fontupper=\tiny]
\begin{verbatim}
---Role---
You are a helpful assistant responding to questions about data in the tables provided.

{"Single paragraph"}
Answer in {language}
Answer that you do not have data to handle the user questions.
\end{verbatim}
\end{tcolorbox}

\begin{tcolorbox}[colback=black!5!white, colframe=black!50!yellow, title=LLM-only Generation Prompt, fontupper=\tiny]
\begin{verbatim}
---Role---
You answer questions. Respond in {language}.

---Goal---
Give a direct answer in {language}.
If unsure, say you don’t know. Never invent information.

---Target response length and format---
Reply EXTREMELY BRIEFLY in {language}.
Provide only the answer.
\end{verbatim}
\end{tcolorbox}

\begin{tcolorbox}[colback=black!5!white, colframe=black!50!white, title=QA Pairs Translation Prompt, fontupper=\tiny]
\begin{verbatim}
--- Goal ---
Translate the provided question–answer pair into {language}.

--- Task ---
You are a rigorous translator tasked with translating the provided question–answer pair into {language}.

--- Rules ---
- Translate the given question and answer into {language}.
- Do not answer the translated question in any way.
- Do not add anything to the translated question or answer.
- Do not add quotation marks to the translated question or answer.
- Keep intact any names, titles, etc. that are in a foreign language.

--- Important ---
WHILE YOU TRANSLATE INTO {language}, DO NOT ANSWER THE QUESTION IN ANY WAY!
                  
--- Output Format ---
Respond only with the following format:
Question: <question translated into {language}>
Answer: <answer translated into {language}>

--- Input ---
Question: {question}
Answer: {answer}
\end{verbatim}
\end{tcolorbox}

\begin{tcolorbox}[colback=black!5!white, colframe=black!50!white, title=LLM-as-judge Evaluation Prompt, fontupper=\tiny]
\begin{verbatim}
--- Goal ---
Evaluate the quality of a response generated by a RAG in a question answering task.

--- Task ---
You are a rigorous evaluator tasked with assigning a score from 0 to 10 to the RAG response.

--- Criteria ---
- Relevance of the response: Does the RAG response directly address the user’s question?
- Faithfulness to the expected answer: Does the RAG response match the reference answer without adding invented or contradictory
information?

--- Rules ---
- If the RAG response is absent or states that the information is not available, assign 0.
- If the RAG response is off-topic or not relevant to the question, assign a low score.
- If the RAG response contradicts the expected answer or invents information, assign a low score.
- If the RAG response suggests that the documents do not provide any useful information for the question, assign a low score.
- If the RAG response is correct, complete, sourced on the documents, and well aligned with the expected answer, assign a
high score.

--- Output Format ---
Answer using only the following format:
Score: <number from 0 to 10>  
Justification: <brief explanation of the score>

--- Input ---
Question: {question}
Expected Answer: {expected}
RAG Response: {generated}
\end{verbatim}
\end{tcolorbox}

%%%%%%%%%%%%%%%%%%%%%%%%%%%%%%%%%%%%%%%%%%%%%%%%%%%%%%%%%%%%%%%%%%%%%%%%%%%%%%%
%%%%%%%%%%%%%%%%%%%%%%%%%%%%%%%%%%%%%%%%%%%%%%%%%%%%%%%%%%%%%%%%%%%%%%%%%%%%%%%

\end{document}